# Automated Detection of Cat Facial Landmarks

George Martvel[1] · Ilan Shimshoni[1] · Anna Zamansky[1]

## Abstract

The field of animal affective computing is rapidly emerging, and analysis of facial expressions is a crucial aspect. One of the most significant challenges that researchers in the field currently face is the scarcity of high-quality, comprehensive datasets that allow the development of models for facial expressions analysis. One of the possible approaches is the utilisation of facial landmarks, which has been shown for humans and animals. In this paper we present a novel dataset of cat facial images annotated with bounding boxes and 48 facial landmarks grounded in cat facial anatomy. We also introduce a landmark detection convolution neural network-based model which uses a magnifying ensemble method. Our model shows excellent performance on cat faces and is generalizable to human and other animals facial landmark detection.

**Keywords** Landmarks · Detection · Ensemble models

## 1 Introduction

There is a huge body of work addressing automated human facial analysis, which has a plethora of applications in affective computing, healthcare, biometry, human-computer interaction and many other fields (Friesen & Ekman, 1978; Li & Deng, 2020). One of the cornerstones of most of the approaches is localization of keypoints, also known as fiducial points or facial landmarks (Wu & Ji, 2019). Their location provides important information that can be used for face alignment, feature extraction, facial expression recognition, head pose estimation, eye gaze tracking and many more tasks (Akinyelu & Blignaut, 2022; Al-Eidan et al., 2020; Malek & Rossi, 2021). Automated facial action unit recognition systems also often use facial landmarks (Yang et al., 2019; Ma et al., 2021).

The type and number of landmarks varies depending on the specific application at hand: for coarse tasks such as

Communicated by Helge Rhodin.

✉ George Martvel
  martvelge@gmail.com

  Ilan Shimshoni
  ishimshoni@is.haifa.ac.il

  Anna Zamansky
  annazam@is.haifa.ac.il

[1] Information Systems Department, University of Haifa, Haifa, Israel

face recognition or gaze direction assessment a few landmarks may suffice, however for facial expression analysis, depending on the species, up to eighty landmarks may be needed, leading to the need for automation of landmark detection (Tarnowski et al., 2017; Wu et al., 2018). Despite its being seemingly a simple task, this computer vision problem has proven to be extremely challenging due to inherent face variability as well as effects of pose, expression, illumination, etc. Accordingly, a huge body of research addresses the automation of facial landmark detection and localization.

Research concerned with automation of animal behavior analysis has so far lagged behind the human domain. However, this is beginning to change, partly due to introduction of deep-learning platforms such as the DeepLabCut (Mathis et al., 2018), DeepPoseKit (Graving et al., 2019), which allow to automate animal motion tracking and recognition of keypoints located on animals' body. However, recently an increasing number of works go 'deeper' than tracking addressing recognition of animals' affective states, including emotions, stress and pain. A comprehensive survey of state-of-the-art of these methods is provided in Broome et al. (2023), focusing mainly on facial expressions. Since they are produced by all mammals, they are one of the most important channels of communication in this context (Hummel et al., 2020; Paul & Mendl, 2018). Facial analysis of animals presents immense challenges due to the great variability of textures, shapes, breeds and morphological structure in the realm of animals. This leads to the need for addressing the problem of automated facial landmark detection in animals.

This brings about unique challenges not present in the human domain, such as more complicated data collection protocols and ground truth establishment in the absence of verbal communication with animals. But most importantly, the variabilities across and within lead to the need of exploring effective and economic, in terms of computational resources and time, approaches for solving the landmark localization problem, taking into account the lack of the proper amount of data (Broome et al., 2023) needed to solve such a problem by machine learning methods.

To address these gaps, this paper makes two main contributions. First, we introduce the Cat Facial Landmarks in the Wild (CatFLW) dataset containing 2091 images of cats with 48 facial landmarks, annotated using an AI-assisted annotation method and grounded with the cat facial anatomy. Secondly, we present an Ensemble Landmark Detector (ELD), a baseline model which shows excellent performance on cats. The model also performs well on the human facial landmark WFLW dataset, compared to state-of-the-art models, as well as scaling the problem of detecting facial landmarks to other animal species.

## 2 Related Works

### 2.1 AI-assisted Annotation

AI-assisted annotation of training data is widely used to reduce the time and cost of created datasets for machine learning (Wu et al., 2022). The idea of the method is to consistently annotate training data using predictions of a machine learning model that is gradually being retrained on corrected data. Due to the gradual improvement of the predictions of the assisting model and the use of algorithms and metrics for selecting the most suitable training data, human annotators spend less time processing more data.

Various AI-assisted annotation techniques are also used in computer vision tasks: image and video classification (Li & Guo, 2013; Collins et al., 2008; Yang et al., 2003; Gu et al., 2015), object detection (Yoo & Kweon, 2019; Liu et al., 2021; Aghdam et al., 2019; Kellenberger et al., 2019), segmentation (Sinha et al., 2019), face recognition (Elhamifar et al., 2013) and landmark detection (Quan et al., 2022).

In the field of animal landmark detection, similar methods are used to speed up the annotation process. So, in LEAP (Pereira et al., 2019), the structure of which includes the gradual training of models for detecting animal poses, Pereira et al. annotated 32 landmarks on *drosophila*. The authors demonstrate that in six iterations of AI-assisted annotation, the speed of annotating one image increased by 20 times. A similar pipeline is also used in Graving et al. (2019), in which the authors created a DeepPoseKit—a universal tool for detecting the poses of insects and animals. Its functionality includes an optimized process of active learning, which has made it a popular tool among the scientific community.

### 2.2 Landmarks in Animals

Facial and body landmarks are increasingly being used to assess various internal states of animals. Indeed, by measuring geometric relationships and analyzing the position of animal body parts, it is possible to establish correlations between the internal states of animals and their external expressions (Brown et al., 2013; Bierbach et al., 2017; Kain et al., 2013; Wiltschko et al., 2015).

Finlayson et al. (2016) evaluated the influence of the types of interaction with mice (*Rattus norvegicus*) on the change in facial expressions. To do this, they processed 3,000 frames, where, among other metrics, measured the width and height of the eye and eyebrow, and angles of the ear and eyebrow, relying on the Rat Grimace Scale (Sotocinal et al., 2011) in the choice of metrics. According to our estimates, to compute these metrics the authors would require 14 symmetrical facial landmarks.

In Ferres et al. (2022), classify emotional states of dogs using data on their postures. The authors trained a DeepLabCut (Mathis et al., 2018) framework model on 13,809 dog images annotated with 24 body landmarks. The article does not provide the accuracy of the landmark detection, however, using them, four classifiers with different architectures trained on 360 instances of body landmarks coordinates demonstrated classification accuracy above 62%.

In McLennan and Mahmoud (2019), propose to combine Sheep Pain Facial Expression Scale (SPFES) (McLennan et al., 2016) with landmark-based computer vision models to evaluate pain in sheep. However, the authors emphasise that "availability of more labelled training datasets is key for future development in this area".

Gong et al. (2022) used 16 body landmarks to further classify poses in cows. Using automatically detected landmarks the authors achieve more than 90% precision rate on pose classification.

Often facial landmarks are used for animals for facial alignment, and then other computer vision models use aligned images for animal identification or other purposes (Clapham et al., 2022; Billah et al., 2022). This approach does not require much computational power, since it is limited to a small number of landmarks, but it significantly simplifies the processing for subsequent models, since it normalizes the data.

### 2.3 Cat Facial Landmarks Applications

Many studies (Bennett et al., 2017; Humphrey et al., 2020; Scott & Florkiewicz, 2023) related to the analysis of the internal state of cats have been inspired or directly use Facial Action Coding System for cats (CatFACS) (Caeiro et al., 2017). In Deputte et al. (2021), studied more than 100 h of cat-cat and cat-human interactions to study the positions of ears and tails during these interactions. The authors performed the entire analysis manually without using any automation, which led to significant time and labour costs. In the limitations section, the authors indicate that "a much larger number of data should have been obtained". In Llewelyn and Kiddie (2022) explore the use of CatFACS as a welfare assessment tool for cats with cerebellar hypoplasia (CH), finding 16 action units, which could infer the welfare of healthy and CH cats.

Cats are of specific interest in the context of pain, as they are one of the most challenging species in terms of pain assessment and management due to a reduced physiological tolerance and adverse effects to common veterinary analgesics (Lascelles & Robertson, 2010), a lack of strong consensus over key behavioural pain indicators (Merola & Mills, 2016) and human limitations in accurately interpreting feline facial expressions (Dawson et al., 2019). Three different manual pain assessment scales have been developed and validated for domestic cats: the UNESP-Botucatu multidimensional composite pain scale (Brondani et al., 2013), the Glasgow composite measure pain scale (CMPS) (Reid et al., 2017), and the Feline Grimace Scale (FGS) (Evangelista et al., 2019). The latter was further used for a comparative study in which human's assignment of FGS to cats during real time observations and then subsequent FGS scoring of the same cats from still images were compared. It was shown that there was no significant difference between the scoring methods (Evangelista et al., 2020), indicating that facial images can be a reliable medium from which to assess pain.

In Evangelista et al. (2019), used 8 different metrics on cats' faces in order to establish the connection of facial movements with pain. Those metrics included distances between the ear tips and ear bases, eye height and width, muzzle height and width, as well as two ear angles. In order to measure these parameters, the authors had to manually annotate at least 24 facial landmarks for 51 images. According to their results, cats in pain had a reduced size of muzzle and ears more flattened.

In Yang and Sinnott (2023), used different deep learning approaches to classify pain in cats. Despite the high accuracy of the classification, their work lacks the explicability of the results obtained. The approach based on landmarks allows implementing action units into the procedure of pain or emotion recognition, which makes it more explainable, in contrast to the "black box" deep learning approach. In Feighelstein et al. (2023), 48 facial landmarks were used to identify important facial areas for the same task of pain classification. The presented results clearly demonstrate the most significant aspects by which pain can be determined in cats, and, therefore, can provide additional insights in understanding cats behaviour and their well-being. While such explainability is not critically important for humans, where we can fully rely on our understanding of the expressions of human emotions and on the direct feedback, this is not so for animals.

Finka et al. (2019) applied 48 geometric landmarks to identify and quantify facial shape change associated with pain in cats. These landmarks were based on both the anatomy of cat facial musculature, and the range of facial expressions generated as a result of facial action units (Caeiro et al., 2017). The authors manually annotated the landmarks, and used statistical methods (PCA analysis) to establish a relationship between PC scores and a validated measure of pain in cats. Feighelstein et al. (2022) used the dataset and landmark annotations from Finka et al. (2019) to automate pain recognition in cats using machine learning models based on facial landmarks grouped into multivectors, reaching accuracy of above 72%. This indicates that the use of the 48 landmarks from Finka et al. (2019) can be a reliable method for pain recognition from cat facial images.

Holden et al. (2014) annotated 59 images with 78 facial landmarks and 80 distances between them to classify pain in cats. The authors claim that using facial landmarks and specific distance ratios, it is possible to determine pain in cats with an accuracy of 98%. It should be pointed out that not all the landmarks turned to be useful for determining pain, and collecting data on a larger number of annotated images could provide additional insights about pain markers, for example, dividing the studied cats by affecting medications in order to take into account their effects on the animal's body.

In Vojtkovská et al. (2020), state that the analysis of action units and grimace scales can serve as an accurate way to determine pain in cats (Evangelista et al., 2019; Caeiro et al., 2017), however "screenshots are obtained from videos, but their analysis is time-consuming. In practice, pain should be assessed immediately and easily (Evangelista et al., 2020)".

As can be seen, facial and body landmarks are widely used to analyze the external manifestations of the internal state of animals. For cats, such an analysis is especially important due to the morphological structure and specifics of their reaction to medications. The connection of facial landmarks with the morphological structure of cats' faces allows linking geometric relations to action units and FACS. Creating a tool that allows automatic detection of facial landmarks for cats would significantly speed up and simplify the data collection process for the above-mentioned studies. Our study represents the first of its kind combination of a dataset of cat facial landmarks and a model that allows detecting them with high accuracy for the most anatomically complete display of morphological features of domestic cats.

## 3 The CatFLW Dataset

To promote the development of automated facial landmark detectors in cats, we have created the CatFLW dataset inspired by existing ones for humans and animals. The main motivation for its creation was the relatively small amount of similar datasets or a low number of facial landmarks in existing ones.

### 3.1 Related Datasets

Our dataset is based on the original dataset collected by Zhang et al. (2008), that contains 10,000 images of cats annotated with 9 facial landmarks collected from flickr.com. It includes a wide variety of different cat individuals of different breeds, ages and sexes in different conditions, which can provide good generalization when training computer vision models, however some images from it depict several animals, have visual interference in front of animal faces or cropped (according to our estimates 10–15%). Figure 1 shows the examples of such inapplicable images. It is also worth noting that a number of images contain inaccurate annotations with significant errors, which can also lead to incorrect operation of computer vision models trained on this data.

In Zhang et al. (2008) 9 landmarks are labeled for each image (two for the eyes, one for the nose and three for the each ear), which is sufficient for detecting and analyzing general information about animal faces (tilt of the head or direction of movement), but is not enough for analyzing complex movements of facial muscles.

Sun and Murata (2020) used the same dataset in their work, expanding the annotation to 15 facial landmarks. It slightly better displays the features of the cats' faces structure, containing two more landmarks for each eye, one landmark for the bridge of the nose and one for the mouth in addition to the 9 landmarks presented in Zhang et al. (2008). However, to the best of our knowledge, only 1,706 out of the declared 10,000 images are publicly available.

Khan et al. (2020) collected the AnimalWeb dataset, consisting of an impressive number of 21,900 images annotated with 9 landmarks. Despite the wide range of represented species, only approximately 450 images can be attributed to feline species. Moreover, the annotation system was chosen differently: here there are two landmarks for each eye, one for the nose and four for the mouth.

In Hewitt and Mahmoud (2019), the authors collected a dataset of 850 images of sheep and annotated them with 25 facial landmarks. In our opinion, such a representation is the best in terms of the number of landmarks and their potential application at the moment.

Other relevant datasets, containing less than 9 facial landmarks, are mentioned in Table 1, including those in Liu et al. (2012), Cao et al. (2019), Mougeot et al. (2019), Yang et al. (2016). For comparison, popular datasets for human facial landmark detection (Belhumeur et al., 2013; Le et al., 2012) have several dozens of landmarks. There are also other datasets suitable for face detection and recognition of various animal species (Deb et al., 2018; Guo et al., 2020; Körschens et al., 2018; Chen et al., 2020), but they have no facial landmark annotations.

Two conclusions follow from the above: firstly, the validity of the assumption about the lack of datasets with a comprehensive number of facial landmarks for animals and cats in particular, and secondly, there are differences in the annotations between the datasets and in the choice of landmark positions in accordance with the morphological structure of the animal faces.

### 3.2 The Annotation Process

The CatFLW dataset consists of 2091 images selected from the dataset in Zhang et al. (2008) using the following inclusion criterion which optimize the training of landmark detection models: the image should contain a **single fully visible** cat face, where the cat is in non-laboratory conditions ('in the wild'). We did not put constraints on the breeds and scale of cats and their faces within our guidelines when choosing images in order to maximize the diversification of our dataset. The image sizes range from $240 \times 180$ to $1024 \times 1024$. Figure 2 shows examples of images which were used in the dataset.

We should also consider the issue of partial occlusion of faces, where not all the landmarks are visible, but their location can be estimated from the logic of the structure of the face. Such images, however, were discarded due to considerations of working with convolution neural networks, which largely focus on the visual boundaries and textures of images, as a result of which, when these models are trained on images with incompletely visible faces, bias and incorrect behavior may occur during further predictions. As the practice of working with such models shows, subsequently, when predicting on images where the face is partially occluded, the model nevertheless "guesses" the positions of the landmarks based on the geometric relationship between them.

After selection and filtering of images, a bounding box and 48 facial landmarks were placed on each image using the Labelbox platform (Labelbox, 2023).

**Bounding box.** Each bounding box is defined by the two coordinates of the upper left and lower right corners in the coordinate system of the image. In the current dataset, it is not tied to facial landmarks (this is usually done by selecting the outermost landmarks and calculating the difference in their coordinates), but is visually placed so that the entire face of the animal fits into the bounding box as well as about 10% of the space around the face. This margin is made because when studying detection models on the initial version of the dataset, it was noticed that some of them tend to crop the faces and choose smaller bounding boxes. In the case when face detection is performed for the further localization of facial landmarks, the clipping of faces can lead to the disappearance from the image of important parts of the cat's face, such as the tips of the ears or the mouth. Such an annotation still makes it possible to construct bounding boxes by the outermost landmarks if needed.

**Fig. 1** Examples of non-suitable images from (Zhang et al., 2008). Left to right: ear tip is out of the image, multiple cats, lower part ot the face is occluded

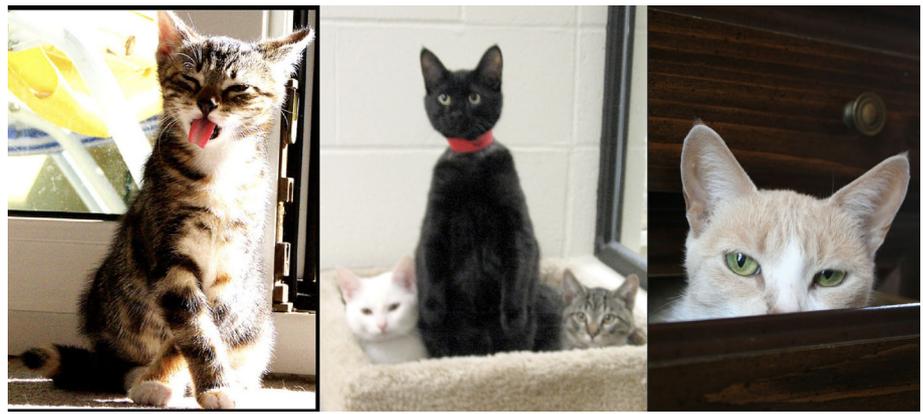

**Table 1** Comparison of animal facial landmarks datasets

| Dataset | Animal | Size | Facial landmarks |
| --- | --- | --- | --- |
| Khan et al. (2020) | Various | **21,900** | 9 |
| Zhang et al. (2008) | Cat | 10,000 | 9 |
| Liu et al. (2012) | Dog | 8351 | 8 |
| Cao et al. (2019) | Various | 5517 | 5 |
| Mougeot et al. (2019) | Dog | 3148 | 3 |
| Sun and Murata (2020) | Cat | 1706 | 15 |
| Hewitt and Mahmoud (2019) | Sheep | 850 | 25 |
| Yang et al. (2016) | Sheep | 600 | 8 |
| Ours (CatFLW) | Cat | 2091 | **48** |

**Fig. 2** Images with bounding boxes and 48 facial landmarks from CatFLW

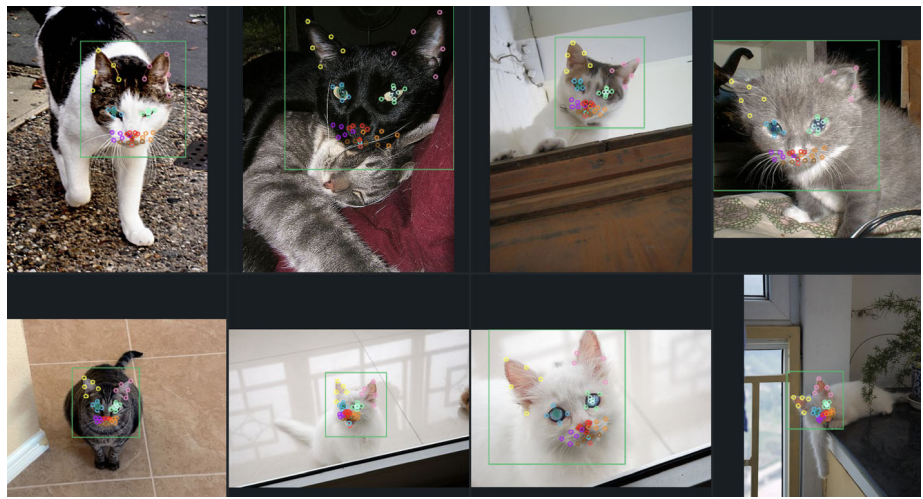

**Landmarks.** 48 facial landmarks introduced in Finka et al. (2019) and used in Feighelstein et al. (2022) were manually placed on each image (shown in Fig. 3). These landmarks were specifically chosen for their relationship with underlying musculature and anatomical features, and relevance to CatFACS (Caeiro et al., 2017). The landmarks are grouped into four semantic groups in the sense of their physical loca-tion on the face: Lower Face (cheeks, whiskers and nose), Jaw (chin and mouth), Upper Face (eyes), and Ears. For each of them, the position guideline and, if possible, the associated facial muscles and action units are determined. Thus, the structure of facial landmarks is quite strictly tied to the anatomical features of the cat's face, which allows associating morphological characteristics with landmarks, their mutual position and movement. Detailed information about the landmarks' placement can be found in the appendices to the article (Finka et al., 2019).

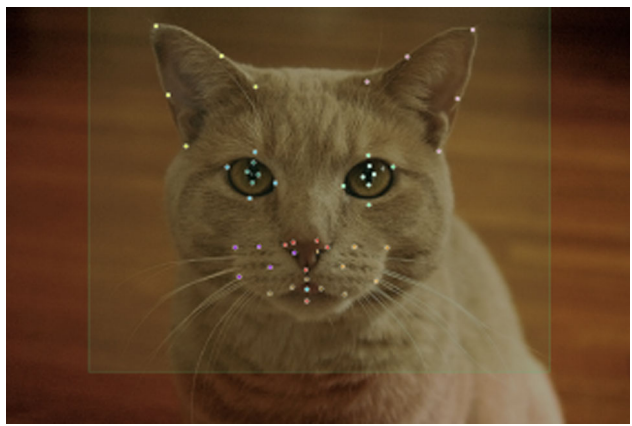

**Fig. 3** Example image from the CatFLW dataset with a bounding box and 48 facial landmarks. The brightness of the image is reduced for greater visibility of facial landmarks

**Annotation Process.** The process of landmark annotation had several stages: at first, 10% images were annotated by an annotator who has an extensive experience labeling facial landmarks for various animals. Then they were annotated by a second expert with the same annotation instructions. The landmarks were then compared to verify the internal validity and reliability via the Inter Class Correlation Coefficient ICC2 (Shrout & Fleiss, 1979), and reached a strong agreement between the annotators with a score of 0.998.

Finally, the remaining part of images were annotated by the first annotator using the "human-in-the-loop" method described below. After the annotation, a review and correction of and landmarks and bounding boxes were performed.

### 3.3 AI-assisted Annotation

The concept behind AI-assisted annotation approach involves the systematic annotation of training data by leveraging the predictions generated by a machine learning model. This model undergoes a continuous retraining process, each time increasing the accuracy and reducing the time required to annotate the consecutive training data.

Khan et al. (2020) spent approximately 5,408 man-hours annotating all the images in the AnimalWeb dataset (each annotation was obtained by taking the median value of the annotations of five or more error-prone volunteers). Roughly, it is 3 min for the annotation of one image (9 landmarks) or 20 s for one landmark (taken per person). Our annotation process took ∼140 h, which is about 4.16 min per image (48 landmarks) and only 5.2 s per landmark. Such performance is achieved due to a semi-supervised annotation method, inspired by such in Graving et al. (2019), Pereira et al. (2019) that uses the predictions of a gradually-trained model as a basis for annotation.

To assess the impact of AI on the annotation process, 3 batches were created from the data. The first one consisted of 200 images annotated without AI assistance and 620 images from Finka et al. (2019) (due to the absence of consent from the authors, we responsibly utilized their dataset solely for pre-training purposes and excluded it from the training data in our subsequent experiments with the ELD model), so 820 images total. Then, using the AI-assisted annotation methodology, we annotated the second batch (910 images) using the ELD model (v1) trained on the first one. The average time for annotation of one image between the first and second batches was reduced by 35%, since for annotation it became necessary only to adjust the position of the landmark, and not to place it from scratch. In the second step of the process, we annotated the third batch (981 images) using predictions of our model (v2) trained on a combination of the first batch and a manually corrected version of the second (∼1700 images total). The results obtained for the time spent on annotation of one image are shown in Fig. 4(left). Increasing the accuracy of predicting the position of a landmark reduces the time required to adjust it, sometimes to zero if the annotator considers the position to be completely correct. Figure 4(right) shows the distribution of the percentage of images in which a specific facial landmark was not shifted by the annotator: between the two versions of the model, the increase is approximately doubled.

## 4 Ensemble Landmark Detector

The problem of regression of facial landmarks is formulated as follows: having an image $\mathbf{x} \in \mathbb{R}^{H \times W \times C}$, it is needed to create a function $f$ (in this case parameterized by a neural network) that maps the space $\mathbb{R}^{H \times W \times C}$ into the space of $K = 48$ landmarks $L^K$: $f(\mathbf{x}) = [\mathbf{y}_1(\mathbf{x}), \dots, \mathbf{y}_K(\mathbf{x})] \in L^K$. Each landmark $\mathbf{y}_i(\mathbf{x})$ is defined by two coordinates in the $\mathbf{x}$ coordinate system.

### 4.1 Architecture

Grishchenko et al. (2020) point out that "using a single regression network for the entire face leads to degraded quality in regions that are perceptually more significant (e.g. lips, eyes)". Indeed, regression models, with correct training and compliance with the bias-variance tradeoff, are able to provide accurate results only with a relatively small dimension of the output vector. In the case of large dimensions, the model tends to "generalize" or average the output vector: this explains the fact that when obscuring a part of the face, hidden landmarks are still determined from the general geometry of the face and relative position with other landmarks, rather than from the image. Moreover, in the case of the human face, the landmarks are relatively "firmly connected" with each other (with the exception of the eyes and the lower part of the face). This leads to the fact that even with an inaccurate prediction of the regression model, the landmarks will be approximately in the right places, giving a relatively small error in general.

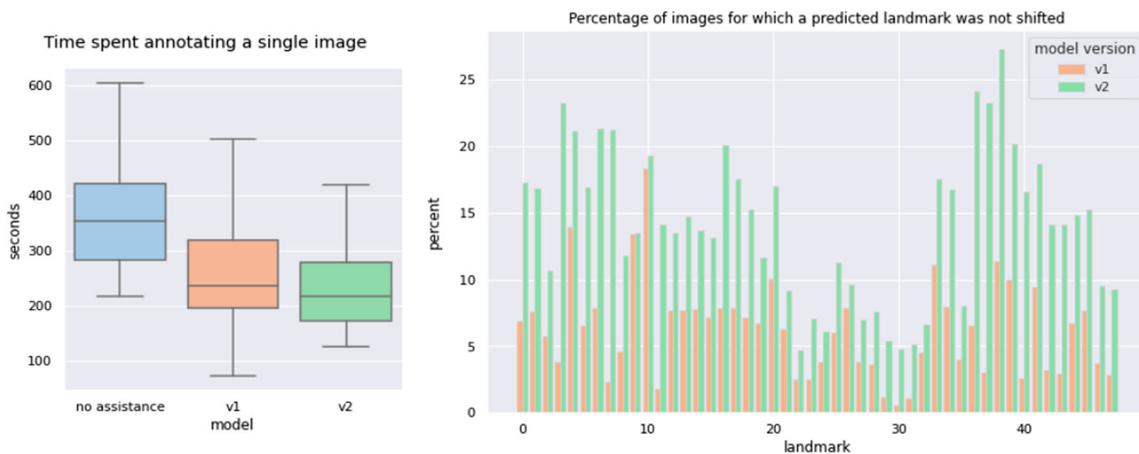

**Fig. 4** Left: The distribution of annotation time per image for different batches. The predictions of the first model (v1) significantly reduced the annotation time, while the subsequent refinement by the second model (v2) reduced the variance, without changing the median value too much. Right: The percentage of images from the batch for which a predicted landmark was not shifted. The more accurate the model's predictions are, the less time is needed for annotation: correctly predicted landmarks do not need to be shifted

In Grishchenko et al. (2020), the authors distinguish three regions of interest on the human face: two eyes and lips, since they have the greatest mobility. By analogy, we will use the magnification of regions to refine landmark coordinates on cats' faces, but in a cascade way. For our model, we will use an ensemble architecture with five regions of interest, since in addition to the eyes and the whiskers area, cats are characterized by high ear mobility.

**Face Detection.** The first stage of landmark detection is localization and cropping of the cat's face from the input image. For this, the image is rescaled with filling to a resolution of $224 \times 224$ and is fed as an input to the face detector. As a detector, we used an EfficientNetV2 (Tan & Le, 2021) model with a custom head. To do this, we disabled the *include_top* parameter when initializing the model and added three fully connected layers with ReLU and linear activation functions and sizes of 128, 64, 4. The model's output is a vector of landmarks with 4 coordinates of the bounding box (upper left and lower right corners). Figure 5 shows the detector's structure.

After detecting and cropping the image by the bounding box, the resulting image is again rescaled to $224 \times 224$ and passed for detection of regions.

**Regions Detection.** In the next step, we used a model similar to the face detector, except that the output layer has size of 10. Thus, we detect the coordinates of 5 centers of regions of interest: eyes, nose (whiskers area) and two ears. We used taking average landmark coordinates from the corresponding regions to get 5 landmarks from 48 as training data.

**Ensemble Detection.** After determining the centers of the regions of interest, the $224 \times 224$ image is aligned by the eyes to reduce the spread on the roll tilt angles and five fixed-size regions are cropped out of it. For the eyes, the region size is set as a quarter of the image (56x56 pixels), for the nose and ears—half of the image ($112 \times 112$ pixels). These sizes were taken empirically so that all associated landmarks were located within the regions in most cases. We choose fixed sizes of regions for greater normalization of the obtained data: in the case of automatic detection of regions' bounding boxes, the data contained in the cropped regions would be much more heterogeneous from the standpoint of the position and cropping of potentially relevant parts of the face (for example, for the eye region, the center of the eye is almost always in the center of the region in our pipeline). The regions are then resized to $224 \times 224$ to meet the input size of the EfficientNetV2 model.

Then the landmarks are divided into groups according to their corresponding regions. There are 8 landmarks for each eye, 5 for each ear, and 22 for the nose and whiskers region. All landmarks are then detected by an ensemble of five models, each of which has a structure similar to the detector, except that the size of the output layer is equal to the number of landmarks multiplied by two (the coordinates of the corresponding landmarks). After that, the coordinates of the landmarks are transferred to the coordinate system of the original image and are combined into a final vector of 96 coordinates (48 landmarks). The final structure of the Ensemble Landmark Detector is shown in Fig. 6.

**Fig. 5** EfficientNetV2-based detector structure. The model takes a 224 × 224 RGB image and outputs the coordinates vector (top left and bottom right corners of the bounding box)

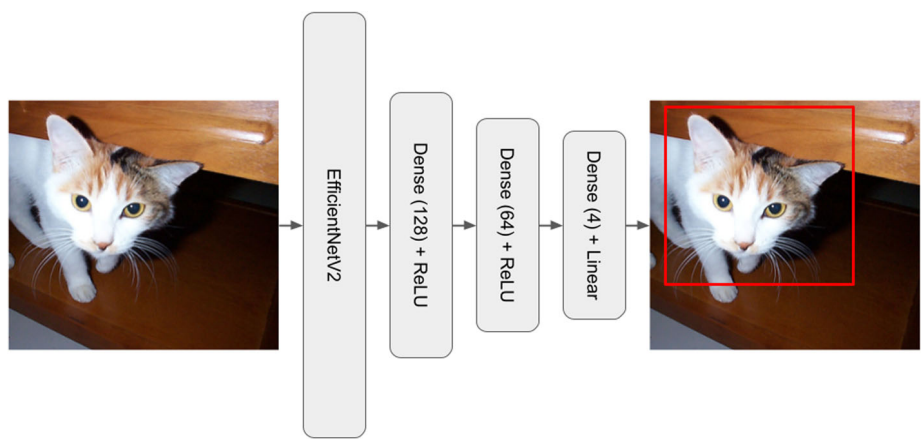

## 5 Experiments

We present an ablation study of various versions of our model on CatFLW in order to (1) provide a baseline for detecting landmarks on our dataset, and (2) demonstrate the effectiveness of the proposed features of the ELD model.

To evaluate the accuracy of the models, we will use the commonly used Normalized Mean Error (NME) metric, which is defined as

$$NME = \frac{1}{M \cdot N} \sum_{i=1}^{N} \sum_{j=1}^{M} \frac{\left\| x_i^{\,j} - x_i'^{\,j} \right\|}{iod_i},$$

where $M$ is the number of landmarks in the image, $N$ is the number of images in the dataset, $iod_i$ is the inter-ocular distance (distance between the outer corners of the two eyes), $x_i^{\,j}$ and $x_i'^{\,j}$—the coordinates of the predicted and the ground truth landmark respectively.

### 5.1 Experimental Setup

Since all parts of ELD are based on the same backbones, we used the identical training strategy for the face detector, the detector of the centers of regions of interest and each of the five models in the ensemble. We trained our models for 300 epochs using the mean squared error loss and the ADAM optimizer with a starting learning rate of $10^{-4}$ and a batch size of 16. We lowered the learning rate by a factor of ten each time the validation loss did not improve for 75 epochs, saving a model with the best validation loss.

We used augmentation of the training data, artificially doubling the size of the training dataset, applying each of the following methods with a 90% probability to each pair of image-landmarks: random rotation, changing the balance of color, brightness, contrast, sharpness, laying-on blur masks and random noise.

Our hardware consisted of a Supermicro 5039AD-I workstation with a single Intel Core i7–7800X CPU (6 cores, 3.5GHz, 8.25M cache, LGA2066), 64GB of RAM, a 500GB SSD, and an NVIDIA GP102GL GPU.

### 5.2 Ablation Study

To evaluate the accuracy of our model on the CatFLW, we split the dataset into training, validation and test parts in the ratio of 75:15:10 (1569/314/208 images correspondingly). All experiments with the dataset were carried out using the subsets obtained from this division.

**Face Detection.** Our goal in developing the CatFLW dataset and ELD model was to make an end-to-end system capable of working in the wild, that is, on unannotated images. Due to the fact that most of the models for detecting facial landmarks do not focus on detecting faces (Sun et al., 2019; Wang et al., 2019; Li et al., 2020, 2022; Jin et al., 2021; Huang et al., 2021; Lan et al., 2021; Zhou et al., 2023),but use ready-made bounding boxes at the data pre-processing stage, we do not deviate from this pipeline in this section later for the fare comparison. However, we provide data for detecting landmarks in the case of using our manually labeled bounding boxes and bounding boxes determined by the ELD facial detector. In the experiments we used a series of EfficientNetV2 models (Tan & Le, 2021) as backbones. For additional comparison, we also trained YOLOv8 [86], using the YOLOv8n version, trained for 300 epochs with the default hyperparameters.

**Fig. 6** Ensemble Landmark Detector. Provided ensemble architecture and region centers could be changed depending on the face morphology

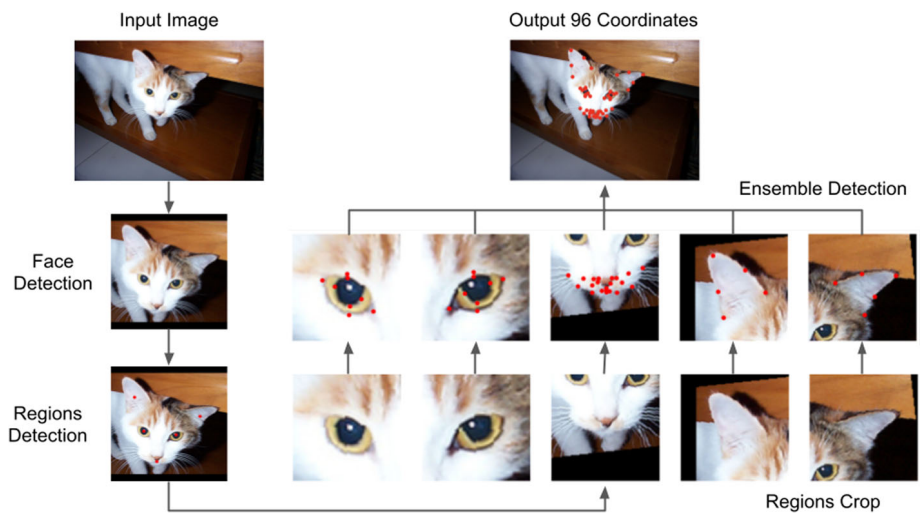

**Table 2** Comparison of landmark detection using training data obtained by facial detector and with pre-cropped faces on the CatFLW test set

| Model | NME |
| --- | --- |
| – | 2.98 |
| YOLOv8n (Jocher et al., 2023) | 4.08 |
| EfficientNetV2B3 (Tan & Le, 2021) | 4.45 |
| EfficientNetV2B2 (Tan & Le, 2021) | 4.69 |
| EfficientNetV2B1 (Tan & Le, 2021) | 4.76 |
| EfficientNetV2B0 (Tan & Le, 2021) | 5.09 |

In all cases, the EfficientNetV2B0 backbone was used in ensemble detection

**Table 3** Comparison of landmark detection using different regions' of interest sizes proportions in relation to the face bounding box

| Region Varying | Size | NME |
| --- | --- | --- |
| Eyes | 1/2 | 3.11 |
| | 1/3 | 3.07 |
| | **1/4** | **2.98** |
| | 1/5 | 3.02 |
| | 1/6 | 3.08 |
| Nose | 1 | 3.31 |
| | **1/2** | **2.98** |
| | 1/3 | 3.23 |
| Ears | 1 | 3.22 |
| | **1/2** | **2.98** |
| | 1/3 | 3.21 |

In all cases, the EfficientNetV2B0 backbone was used

From the direct comparison given in Table 2, we can say that when using a facial detector, the accuracy drops significantly. This is explained by the fact that in the case of pre-cropped faces, the scale and approximate position of the landmarks in the image are preserved. In the case of a detector, there may be variations in the scale of the detected faces, as well as "cutting-off" meaningfully important facial parts potentially containing landmarks.

In the case of the YOLO model, we observe a higher final accuracy of landmark detection, however, such an improvement leads to a disadvantage—in some cases, the model does not detect face bounding boxes, which leads to the inability to detect landmarks in such images. For our test set, we didn't get landmarks in five images (2.3% of the total). For comparison, this result is presented here without taking into account the missing images, however, we do not use the YOLO detector in the ELD to ensure the experiments validity.

**Regions of Interest.** When choosing the size of the regions for the ensemble detection, we used empirical values based on the considerations that for too small regions, some of the information will be cut off and the coordinates of the desired landmarks will lie outside the region of interest, and for too large regions, there will be a lot of extra information in each, which ultimately reduces the accuracy of detection.

To test our assumptions, we trained ensemble models on regions of different sizes, changing one region size for each experiment (by default, we consider the same region sizes for pairs of eyes and ears). For convenience, we took the sizes of the regions as a proportional part of the rescaled face bounding box ($224 \times 224$ pixels). The final NME metric is given for all the landmarks, in order to measure the contribution of changes in each region in total. The results confirming our suppositions are shown in Table 3. Similar experiments were conducted for several types of backbones and showed analogous dependencies.

**Backbones.** The EfficientNet models were chosen as a backbone for the ELD because of the balance between the number of parameters and the accuracy of performance. Since such backbones are rarely used for regression problems on their own, we could only refer to the accuracy on image classification datasets (Xie et al., 2017; Tan & Le, 2019; Liu et al., 2022) and to the use of these models as backbones in other landmark detection models (Mathis et al., 2018, 2021). For the experiment, we have selected several popular and most effective backbones and measured the accuracy of the prediction of the ELD model as a whole, where each of its components is a certain backbone. The results are shown in the Table 4.

**Table 4** Evaluation of a total landmark detection error on the CatFLW using different backbones

| Backbone | Param. (M) | NME |
|---|---|---|
| MobileNetV2 (Sandler et al., 2018) | 3.5 | 3.09 |
| EfficientNetB0 (Tan & Le, 2019) | 5.3 | 2.97 |
| EfficientNetV2B0 (Tan & Le, 2021) | 7.2 | 2.98 |
| EfficientNetB1 (Tan & Le, 2019) | 7.9 | 2.98 |
| EfficientNetV2B1 (Tan & Le, 2021) | 8.2 | 2.96 |
| EfficientNetB2 (Tan & Le, 2019) | 9.2 | 2.96 |
| EfficientNetV2B2 (Tan & Le, 2021) | 10.2 | 2.93 |
| EfficientNetB3 (Tan & Le, 2019) | 12.3 | 2.94 |
| DenseNet169 (Huang et al., 2017) | 14.3 | 2.92 |
| EfficientNetV2B3 (Tan & Le, 2021) | 14.5 | 2.91 |
| EfficientNetV2S (Tan & Le, 2021) | 21.6 | 2.83 |
| ConvNeXtTiny (Liu et al., 2022) | 28.6 | 2.78 |
| ResNet101 (He et al., 2016) | 44.7 | 2.91 |

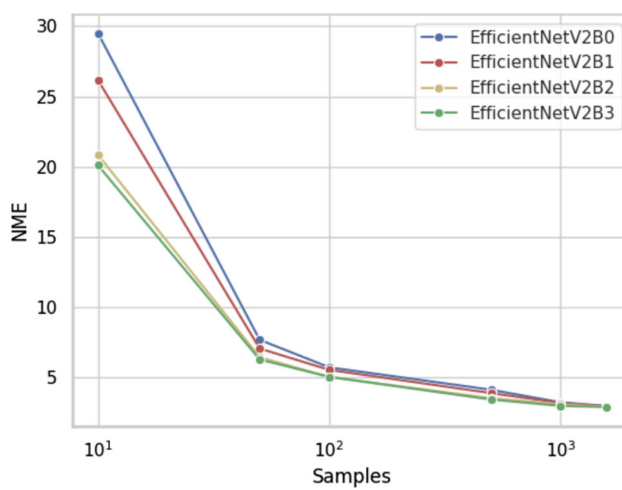

**Fig. 7** ELD's prediction error on the test set with different numbers of training examples

It can be seen that more complex backbones generally show better results (sometimes more architecturally advanced models with fewer parameters demonstrate better accuracy), but a slight increase in accuracy is caused by a significant increase in the number of parameters. Considering that the ELD contains six similar models, we do not provide more results for complex backbones due to their inefficiency for detection.

**Data Size Impact.** For machine learning models, the problem of "data hunger" is well known, that is, a strong dependence of the accuracy of the model on the amount of training data. In Mathis et al. (2018), the authors show that the DeepLabCut model for detecting body landmarks reaches an acceptable error of several pixels, starting with hundreds of annotated images, when the error decreases with an increase in the test set, but much less significantly. A similar correlation can be observed for the DeepPoseKit (Graving et al., 2019) model.

We present a study of the effect of the size of the training set on the accuracy of the ELD (Fig. 7). Our model also demonstrates acceptable results, starting with several hundred images in a training set, although further increase in data size still leads to a noticeable increase in accuracy.

## 5.3 Comparison with Other Models on CatFLW

For a comparative accuracy evaluation of the ELD, we measured the perfomance of the several popular landmark detectors on the CatFLW. We have selected the following models: DeepPoseKit (DPK) (Graving et al., 2019), LEAP (Pereira et al., 2019), DeepLabCut (DLC) (Mathis et al., 2018, 2021; Nath et al., 2019) and Stacked Hourglass (Newell et al., 2016). Despite the fact that these models are usually used as pose detectors (Ye et al., 2022), they are capable of detecting various landmarks and are created as general multifunctional models, including those suitable for facial landmarks detection, and widely used in the literature (Labuguen et al., 2019; Mathis & Mathis, 2020; Zhan et al., 2021).

We trained the models on the CatFLW training set and evaluated them on the test set using the NME metric (in all cases, we used pre-cropped faces with a resolution of $224 \times 224$ according to the CatFLW's bounding boxes). The training process was performed using the DeepPoseKit platform (Graving et al., 2019). All models were trained for 300 epochs with a batch size of 16, mean squared error loss, the ADAM optimizer, and optimal parameters for each model (indicated in the corresponding papers). During preprocessing, we used a similar selection of landmark regions as in the ELD, as well as a similar approach to image and landmarks augmentation. The results of the experiments are shown in Table 5.

From these results, it is evident that our proposed ELD model, even with a fewer parameters demonstrates a superior accuracy compared to other models, and only improves performance, albeit with a significant increase in the number of parameters.

**Table 5** Comparison of landmark detection error on the CatFLW dataset using different detection models

| Model | Backbone | Total Param. (M) | NME |
| --- | --- | --- | --- |
| LEAP (Pereira et al., 2019) | SegNet | 2.6 | 7.02 |
| DLC (Mathis et al., 2021) | MobileNetV2 | 1.5 | 4.27 |
| DPK (Graving et al., 2019) | Stacked DenseNet | 21.9 | 3.97 |
| DLC (Mathis et al., 2018) | ResNet50 | 25.4 | 3.82 |
| DLC (Mathis et al., 2018) | ResNet101 | 44.6 | 3.81 |
| Stacked Hourglass (Newell et al., 2016) | Hourglass (8 stacks) | 24.7 | 3.78 |
| ELD | MobileNetV2 | 21.0 | 3.09 |
| ELD | EfficientNetV2B1 | 49.2 | 2.96 |
| ELD | EfficientNetV2S | 129.6 | 2.83 |

**Table 6** Evaluation of landmark detection on WFLW

| Method | Backbone | NME |
| --- | --- | --- |
| DeCaFa (Dapogny et al., 2019) | U-Net | 4.62 |
| HRNet (Sun et al., 2019) | HRNet-W18 | 4.6 |
| Awing (Wang et al., 2019) | Hourglass | 4.36 |
| DAG (Li et al., 2020) | HRNet-W18 | 4.21 |
| LUVLi (Kumar et al., 2020) | DU-Net | 4.37 |
| PIPNet (Jin et al., 2021) | ResNet-101 | 4.31* |
| ADNet (Huang et al., 2021) | Hourglass | 4.14 |
| HIH (Lan et al., 2021) | Hourglass | 4.08 |
| DTLD (Li et al., 2022) | ResNet-18 | 4.08* |
| RePFormer (Li et al., 2022) | ResNet-101 | 4.11 |
| SPIGA (Prados-Torreblanca et al., 2022) | Hourglass | 4.06 |
| STAR (Zhou et al., 2023) | Hourglass | 4.02 |
| LDEQ (Micaelli et al., 2023) | DEQ | **3.92** |
| ELD | EfficientNetV2B1 | 4.65 |
| ELD | EfficientNetV2B1 | 4.05* |

*Denotes that a subset of landmarks were used in the evaluation

## 5.4 WFLW Dataset

We additionally evaluate our method on the popular human face WFLW (Wider Facial Landmarks n-the-Wild) dataset (Wu et al., 2018). It contains 7,500 training images and 2,500 test images annotated with 98 landmarks.

For ensemble detection, we disable our face detection model, since the bounding boxes for this dataset are commonly preset by landmarks. We do not change the architecture of the model, besides dividing the human face into four regions of interest: eyes, nose and mouth. We separately detect 33 landmarks on the lower part of the face, as they are not suitable for our magnifying method. The same training parameters were used to train the models as for training similar models on CatFLW. We compare ELD with several state-of-the-art methods, as shown in the Table 6.

Some of the results for our model are given without the detection of the landmarks on the lower border of the face. Due to the impossibility of using the magnifying ensemble method, the detection accuracy on jaw landmarks is signif-icantly worse than on the rest. Table 7 shows the NME for landmarks taken on individual parts of the face.

The provided results demonstrate that our model handles localized groups of landmarks better, being significantly inferior in the case of non-localized ones. When detecting a partial set of landmarks, our model demonstrates results comparable to the state-of-the-art.

## 5.5 Complex Cases

To further evaluate the effectiveness of the ELD, we conducted a qualitative analysis of its operation on particularly challenging images that are not presented in the training dataset. By challenging images, we mean images of cats not from the CatFLW, in which their faces are partially occluded, heavily rotated, or there are several cats in the image. We studied the performance of the model on 50 selected images that fit the criteria.

The limitation of the current ELD architecture is the detection of a single object in the image due to the fixed output of the facial detection model (regardless of whether the object is present at all or there are many of them). The solution to this problem can be the use of a YOLOv8-based detector, in which case both the detection of multiple objects and their classification are possible.

**Table 7** NME on different face parts using EfficientNetV2B1 as a backbone

| Left Eye | Right Eye | Nose | Mouth | Jaw |
|---|---|---|---|---|
| 4.0 | 3.98 | 3.54 | 4.37 | 5.83 |

The peculiarity of the ensemble structure is the incoherence of the parts between each other. Therefore, for example, with partial occlusion or extreme position of the cat's face, unlike other models, which in such cases produce "average" predictions, ELD demonstrates a different behavior. For those parts that are visible, landmarks are determined almost precisely, and for occluded ones, relatively random prediction in relation to other landmarks (but still consistent within the region) may occur.

Examples of the ELD's performance on complex cases are shown in Fig. 8.

### 5.6 Transfer to Other Cat Species

To investigate the generalization abilities of our model, we additionally tested it on images of other *felidae* cat species. Due to the lack of annotated facial landmarks for other species, we were able to conduct only a qualitative study of the detection accuracy. During testing, it was noted that due to the ensemble structure of the detector, several factors affect the accuracy of detection.

Firstly, the position of the detected bounding box and the five centers of the regions of interest varied greatly depending on the appearance of the cat. For example, for lions (*panthera leo*), a strong difference in detection was visible for male and female individuals, since the ELD trained on domestic cats often detected the mane as part of the face.

Secondly, the detection accuracy is influenced by specific morphological features of individuals of different families. For example, it was noticed that the *pantherinae* subfamily has a flat lower jaw shape, which is unusual for domestic cats, and therefore negatively affects the detection of the corresponding model in the ensemble. On the other hand, it turned out that the shape of the ears does not greatly affect the accuracy of detection, at least visually for both rounded and pointed ears.

As a result, we can say that the closer the anatomically examined species are to a domestic cat, the more acceptable the detection results are. So, for leopards (*panthera pardus*), jaguars (*panthera onca*), tigers (*panthera tigris*) and lions (*panthera leo*) belonging to a different subfamily than domestic cats (*felis catus*), detection visually works worse in the nose and mouth area. For more closely related specimens, such as cheetah (*acinonyx jubatus*) or lynx (various *lynx* species), detection works more accurately due to greater external similarity. This implies that when training on new data containing various feline species according to the same annotation scheme with 48 landmarks, the ELD will demonstrate great detection accuracy, although here it is worth noting the potential need to create new annotation systems based on the specifics of anatomy for different species. Examples of the ELD's performance on various cats are shown in Fig. 9.

## 6 Conclusions and Future Work

The field of animal affective computing is only just beginning to emerge. One of the most significant obstacles that researchers in this field currently face is the scarcity of high-quality, comprehensive datasets, as highlighted in Broome et al. (2023). The structure of annotations in the proposed dataset allows researchers to rely on the features of the anatomy of the feline face, which makes it possible to use it not only for the detection of facial landmarks, but also for a deeper analysis of the internal state of cats. It is our hope that the contributions of this paper will support the ongoing efforts to advance the field, and specifically automated facial analysis for a variety of species.

From the experiments conducted, it can be seen that on the CatFLW dataset, our ELD model surpasses the existing popular landmark detection models in detection accuracy, on average containing significantly more parameters and, as a result, inferior to them in speed. As noted in Graving et al. (2019), when choosing a model from the available ones, the researcher should focus on the accuracy-speed trade-off depending on the problem they encounter. We emphasize that for studies where landmarks are used to detect the position of an animal in real time and a detection error of a few pixels is insignificant, most of the models we tested could be preferable. In turn, in cases where experiments are carried out for a prolonged period and the time spent on data processing is not critical, while the accuracy of detection, for its part, directly affects further results, our model outperforms others.

**Fig. 8** ELD predictions on complex images. Left to right: multiple cats, partial occlusion, extreme head tilt angle. Images are taken from Unsplash (2023)

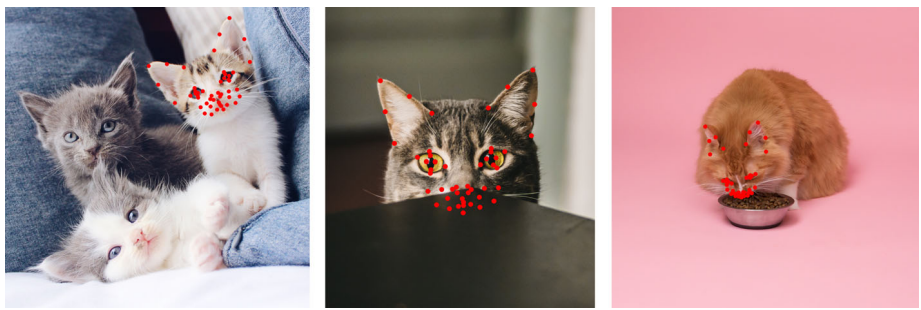

**Fig. 9** ELD predictions on different felidae. Left to right: tiger, cheetah, lion, lynx. Images are taken from Unsplash (2023)

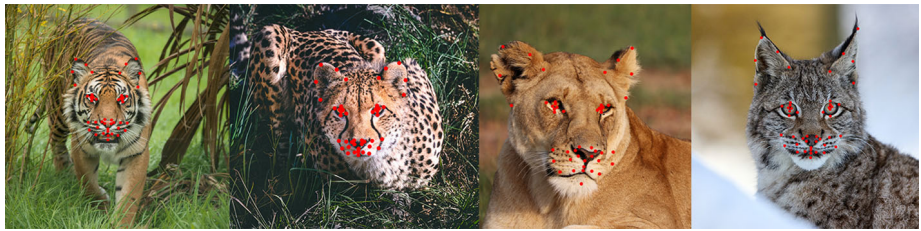

Due to the magnifying method working on semantically grouped landmarks, the ELD model is able to solve the landmark detection problem with high accuracy in a more general sense than only on cat faces, which was demonstrated on a popular human WFLW dataset. The limitation of our approach is the need for a close grouping of landmarks, which is not fulfilled, for example, for jaw landmarks on the same dataset. One of the possible solutions is to further divide the "spaced" landmarks into groups, which, however, may entail a loss of detection efficiency.

An interesting extension of our model can be the replacement of the used face detector with a more "dynamic" YOLOv8. As shown in the Face Detection section, such a detector provides greater detection accuracy on images without preprocessing, although it tends to skip some of them. In cases where this is not critical (for example, when processing video with a large number of frames), such a replacement can provide a high level of detection of landmarks, skipping images in which the cat's face is missing or not fully visible. The study of such an application is one of the directions of our further work on the ELD. Further future research also includes extending the CatFLW dataset with more images and improving the ELD model performance. To do this, we plan to explore new backbones, use new approaches to images augmentation and possible parallelization of computations. We will also investigate how well the model generalizes to other species, such as dogs and monkeys.

We believe that the proposed approaches can serve as a guideline for similar solutions that can solve the problems of identification, detection of emotions and internal states of various animals in an effective and non-invasive way.

**Acknowledgements** The research was supported by the Data Science Research Center at the University of Haifa. We thank Ephantus Kanyugi for his contribution with data annotation and management. We thank Yaron Yossef and Nareed Farhat for their technical support.

**Funding** Open access funding provided by University of Haifa.

**Data availability** The dataset generated during and/or analysed during the current study is available from the corresponding author on reasonable request.